\documentclass[]{llncs}
\usepackage{graphicx}
\usepackage{amsmath,amssymb} 
\usepackage{multirow,url}
\usepackage{color}
\usepackage[width=122mm,left=12mm,paperwidth=146mm,height=193mm,top=12mm,paperheight=217mm]{geometry}
\begin{document}
\mainmatter
\def\ECCV18SubNumber{}  

\title{On Flow Profile Image for Video Representation} 



\author{Mohammadreza Babaee, David Full, Gerhard Rigoll}
\institute{Institute for Human-Machine Communication, TU Munich, Germany}

\maketitle

\begin{abstract}
Video representation is a key challenge in many computer vision applications such as video classification, video captioning, and video surveillance. In this paper, we propose a novel approach for video representation that captures meaningful information including motion and appearance from a sequence of video frames and compacts it into a single image. To this end, we compute the optical flow and use it in a least squares optimization to find a new image, the so-called \textit{Flow Profile Image} (FPI). This image encodes motions as well as foreground appearance information while background information is removed. The quality of this image is validated in activity recognition experiments and the results are compared with other video representation techniques such as dynamic images \cite{8094009} and eigen images \cite{DBLP:journals/corr/abs-1708-05465}.
The experimental results as well as visual quality confirm that FPIs can be successfully used in video processing applications. 

\end{abstract}

\section{Introduction}

Automatic recognition of media data, including handwritten texts~\cite{ehsani2006recognition}, satellite images~\cite{babaee2013assessment,babaee2013immersive}. Due to the importance of visual information, especially for humans, and the ubiquitous presence of cameras in modern society, a large amount of image and video material is constantly being generated.
While images provide valuable appearance-related features about a scene, videos reveal significantly more information.
A video does not only contain more spatial information due to the typically large number of individual frames, but also describes how appearance evolves over time.
However, the main challenge is how to have a compact and informative video representation. 

In this paper, we introduce a novel approach to represent a video sequence as a single image that contains foreground appearance as well as motion information. To achieve this, we utilize optical flow information, computed from RGB frames, and use it in an optimization framework to estimate a single image, the so-called Flow Profile Image (FPI). In particular, to estimate the evolution of the motion intensity in a video, we compute its flow energy profile \cite{DBLP:conf/eccv/WangGZW14} which is a scalar function of time.
, describing the amount of optical flow in each frame.
Next, we determine the FPI such that its projection of the video frames reconstructs the flow energy profile.
In a preprocessing step, the RGB mean of a video is subtracted from all of its frames to remove static background information. 
Applying this technique to several video frames show that this image contains rich source of foreground frame, while redundant background information is removed.

\section{Related work}
\label{sec:relatedwork}
To represent video data well, not only appearance based information has to be captured but also temporal.
Modeling the temporal evolution of appearance makes video data analysis much more difficult than image analysis. While spatial information of individual images can be represented well by convolutional neural networks (ConvNets) with 2D kernels, there is no dominating architecture for video data yet. Encouraged by their tremendous success in image recognition, many video classification approaches based on 2D-ConvNets have been proposed \cite{7558228,SimonyanZ14,7299101}.
Karpathy et al. \cite{6909619} evaluate several methods for extending 2D-ConvNets into video classification.
They investigate four different mechanisms for fusing spatial information across the temporal domain, namely single frame, early fusion, late fusion, and slow fusion. Furthermore, they also explore a multi-resolution ConvNet architecture, consisting of a context stream and a fovea stream, processing the down-sampled original image and the center crop, respectively.
Out of all evaluated techniques, they report the best results for slow fusion.
On fine-tuning top 3 layers of the slow fusion network, classification accuracy was further improved. 
Ng et al. \cite{7299101} explore two principally different architectures based on classic 2D-ConvNets to combine spatial information across longer time periods in videos.
First, they investigate various temporal pooling strategies, namely conv pooling, late pooling, slow pooling, local pooling, and time-domain convolution.
In their experiments, they find that conv pooling works best. 
In a conv pooling architecture, max-pooling is performed after the last convolutional layers across the video frames. 
In a second experiment, the authors model the input video explicitly as an ordered sequence of frames by employing a recurrent neural network with Long Short-Term Memory (LSTM) \cite{DBLP:journals/neco/HochreiterS97} cells. 
By connecting these LSTM cells to the output of a 2D-ConvNet, long-range temporal relationships of the spatial convolutional features can be discovered.
Depending on the actual scenario, sometimes the LSTM based approach performs better and sometimes the conv pooling model.
Donahue et al. \cite{7558228} employ Long Short-Term Memory (LSTM) \cite{DBLP:journals/neco/HochreiterS97} networks to temporally connect spatial features of a 2D-ConvNet for the task of action recognition in videos. 
For the same task, Simonyan and Zisserman \cite{SimonyanZ14} propose a two-stream ConvNet architecture, with one ConvNet being fed by RGB video data to capture spatial information and the other being fed by optical flow (OF) frames to capture information about present motions.
To generate a single prediction, the two streams are combined by averaging or by training a SVM classifier. Due to the great performance benefit, several other two-stream architectures have been proposed  \cite{7780582,8099985,7926610,7298676}. 

As an extension of 2D-ConvNets into time, spatio-temporal 3D-ConvNets can model both spatial and temporal information \cite{8099985,6165309,7410867}. Carreira and Zisserman \cite{8099985} evaluate 3D-ConvNets on the recently published large action classification dataset Kinetics \cite{DBLP:journals/corr/KayCSZHVVGBNSZ17}.
They test different methods for action classification, which consist of an LSTM, a 3D-ConvNet, a two-stream approach, a 3D-fused two-stream method. These approaches are then compared to their proposed new method, Two-Stream Inflated 3D-ConvNets. Even though a 3D-ConvNet is naturally able to capture motion features from pure RGB input videos, they show that their Inflated 3D-ConvNet (I3D) architecture benefits considerably from additional optical flow frames.

To form a single descriptor (representation), all  ConvNet features have to be combined for each video. This is typically done by applying mean-pooling or max-pooling, while a lot of temporal information inherent in video data is lost. To tackle this problem, Fernando et al. \cite{7458903} aggregate features by learning the parameters of a ranking machine.
This method is called rank pooling and is used in several other works  \cite{8094009,7299176,7780700,7780581}.
Moreover, Wang et al. \cite{DBLP:journals/corr/abs-1708-05465} present a method called eigen evolution pooling to summarize a sequence of feature vectors, while preserving as much information as possible.
To do so, the temporal evolution of the respective feature vectors is represented by a set of basis functions that minimize the reconstruction error of the input data.
When applying the previously mentioned pooling methods directly on the RGB pixel intensities of individual video frames, the resulting feature vectors can be interpreted as new images, namely dynamic images \cite{7780700} in case of rank pooling and eigen images \cite{DBLP:journals/corr/abs-1708-05465} in case of eigen evolution pooling.


\section{Approach}
\label{sec:video_representation}

\begin{figure*}[!t]
	\centering
	\def\svgwidth{\columnwidth}
	\includegraphics[width=0.92\textwidth]{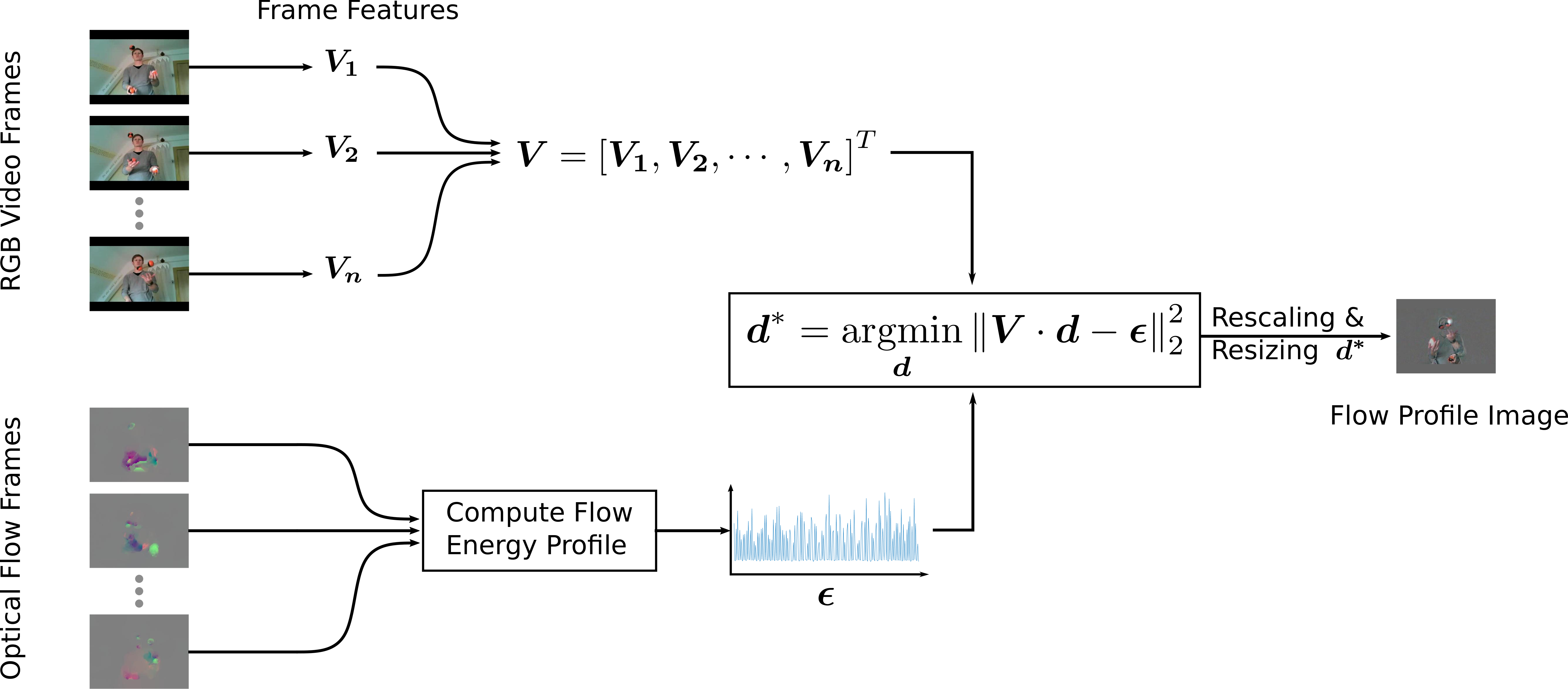}
	\caption{Representation of videos showing the actions (from left to right) blow-drying hair, playing golf, juggling, playing cello, shot putting, typing, hula hooping. Each row corresponds to an image type. From top to bottom: RGB image, flow profile image ($r=1$), flow profile image ($r=2$), dynamic image, and eigen image (first eigen evolution function).}
	\label{fig:flow_profile_images}
\end{figure*}

Inspired by the success of other pooling techniques like rank pooling \cite{7458903} or eigen evolution pooling \cite{DBLP:journals/corr/abs-1708-05465}, we propose a new image pooling method based on the concept of the flow energy profile \cite{DBLP:conf/eccv/WangGZW14} of a video. Our approach fuses a sequence of video frames into a single summarizing image, namely flow profile image. 

\subsection{Flow Energy Profile}
The flow energy profile (FEP) is introduced by Wang et al.~\cite{DBLP:conf/eccv/WangGZW14} for the task of person re-identification. 
It describes how the motion energy intensity evolves over time and is defined as $\epsilon = \left[e_1,e_2,...,e_n\right]^T$ for a sequence of $n$ frames.
We calculate the flow energy of an individual frame $F_i $ as
\begin{equation*}
e_i = \sum_{x,y\in F_i} \left\lVert \left[I_u(x,y),I_v(x,y) \right] \right\rVert_2^2,
\end{equation*}
with $I_u(x,y)$ and $I_v(x,y)$ being the intensity values of its optical flow fields evaluated at the pixel coordinates $\left(x,y\right)$ in the $u$ and $v$ directions,
respectively.
In contrast to the original definition, we do not take the square root when calculating the vector norm to save computation time.
Furthermore, we calculate the flow energy for the whole frame and not only for certain regions, e.g. the legs of a walking person. This enables the generalized usage of this method, without having background knowledge about the video content.
We find that the flow energy profile provides valuable insight about the information content of individual frames. 
When playing tennis, for example, the frames that show a fast moving person with a tennis racket while hitting a tennis ball will have a higher flow energy score than frames showing a person just standing without visible motion.
At the same time, the frame that shows the aforementioned active tennis player is due to the depicted action more discriminative compared to an image of an inactive player.
\subsection{Flow Profile Image}
As in \cite{8094009,DBLP:journals/corr/abs-1708-05465}, we pose the construction of the summarizing images as an optimization problem.
Bilen et al. \cite{8094009} perform rank pooling on the temporally ordered frames of a video and interpret the resulting parameters of the ranking machine as a new image.
In \cite{DBLP:journals/corr/abs-1708-05465}, the authors Wang et al. represent the temporal evolution of RGB features by a set of orthonormal basis functions that minimize the reconstruction error of the input data.
Our basic idea is to find a feature vector $d$ that projects every feature $V_i$ to its corresponding scalar flow energy value $e_i$, i.e. $\forall i: d^T\cdot V_i = e_i$. Note that feature vector $V_i$ is the vectorized frame $F_i$ subtracted by the mean of all frames $\bar{F}$, i.e. $V_i = \text{vec}\left(F_i-\bar{F}\right)$.
The subtraction is done to remove static background information that might hinder the visual encoding of motions in the resulting flow profile images.
Our final vector $d$ encodes data from all features $V_i$, weighted according to the respective flow energy score $e_i$.
Features $V_i$ with a high flow energy value will contribute more to the resulting flow profile image than features with a lower score. 
Since there are in general substantially more parameters in vector $d$ than equations $d^T\cdot V_i = e_i$, the problem can also be treated as a system of linear equation with infinitively many solutions.
One possible solution is performed by computing the pseudoinverse of $V=\left[V_1,V_2,...V_n\right]^T$ and right multiplying the flow energy profile vector, i.e. $d=\text{pinv}\left(V\right)\cdot \epsilon$.
Even though the computation using this method for a single flow profile image is not particularly slow, generating flow profile images for a whole dataset can be quite time consuming.
To reduce the computation time, we calculate an approximate solution for the following optimization problem:

\begin{equation*}
d^* = \underset{d}{\text{argmin}} \underbrace{\left(\frac{\lambda}{2}\cdot\left\lVert d \right\rVert_{2}^2 + \sum_{i=1}^{n}\left\lVert d^T\cdot V_i - e_i\right\rVert_{2}^2\right)}_{=J(d)}
\label{eq:optimization_problem}
\end{equation*}

The first term in $J(d)$ is the typical quadratic regularizer and the second term is the sum of the projection errors for every feature vector $V_i$ to its flow energy value $e_i$.
Inspired by \cite{8094009}, we use the first step of gradient descent as approximate solution for the optimization problem above.
With vector $d_0$ as the starting point and $\eta$ as the initial step size, we obtain

\begin{equation*}
\begin{aligned}
d^* & \approx d_0 - \eta\cdot \nabla J(d)|_{d=d_0} \\
    & \approx d_0 - \eta\cdot \left( \lambda \cdot d_0 + 2 \cdot \sum_{i=1}^{n} \left( d_0^T \cdot V_i - e_i \right) \cdot V_i \right).
\end{aligned}
\end{equation*}

By trying different starting points, including the first frame of the considered video sequence and the frame corresponding to the highest flow energy score, we empirically ascertained that the zero vector as starting point works equally well if not better than others.
At the same time, the whole computation is significantly simplified and reduced to a simple weighted summation of the frame features $V_i$:

\begin{equation*}
\begin{aligned}
d^* & \approx 2\cdot \eta \sum_{i=1}^{n}e_i\cdot V_i \\
   & \propto \sum_{i=1}^{n}e_i\cdot V_i
\end{aligned}
\end{equation*} 
 
To visualize vector $d$, all of its entries have to be in the interval $[0,255]$. Therefore, the proportionality factor $2\cdot \eta$ does not have to be considered when computing the weighted sum.
Instead, we scale all entries of $d$ into the interval $[0,255]$ after summing up the weighted RGB feature vectors $V_i$.

While the flow energy scores can theoretically be used directly, we find that the quality of the resulting images is enhanced when setting the highest $r$ flow energy scores to the same high value and assigning the remaining entries the same low value.
This ensures that not only a single motion cue can be seen well in the resulting images but $r$ number of motion cues.
Assigning a low value to the remaining flow energy scores assures that the generated flow profile image does not become overloaded.

Flow profile images with $r=1$ and $r=2$, dynamic images, and eigen images are shown in figure \ref{fig:flow_profile_images} for various actions.
In the very first row, default RGB frames are shown, each extracted from the middle of a video.
The two rows below comprise flow profile images, the first computed with $r=1$, the second with $r=2$.
Each summarizes the video from which the frames in row one were taken from.
In the fourth row, dynamic images are visualized, computed with the approximate method proposed in \cite{8094009}.
The last row shows eigen images, computed with the first eigen evolution function using our own reimplemented version of eigen evolution pooling.
It can be seen that our flow profile images look especially similar to dynamic images, but also to the eigen images in column two, three, and seven.
But also the eigen images in  have a similar appearance to the respective flow profile images.
When looking closely, it can be observed that flow profile images focus more on specific snippets of motions than dynamic and eigen images do.
The golf player, for example, is shown while swinging his golf club, with individual poses encoded in more detail than in dynamic or eigen images.
Furthermore, the flow profile image with $r=1$ in the last column depicts the woman with the hula hoop at a single characteristic position, while the flow profile image with $r=2$ encodes two poses.
The dynamic and eigen images encode the same action by showing more motion blur.
What all depicted motion snippets in the shown flow profile images have in common, it that they inherit a high motion intensity, determined by the flow energy value in our algorithm.
\begin{figure*}[!t]
	\centering
	\def\svgwidth{\columnwidth}
	\includegraphics[width=0.92\textwidth]{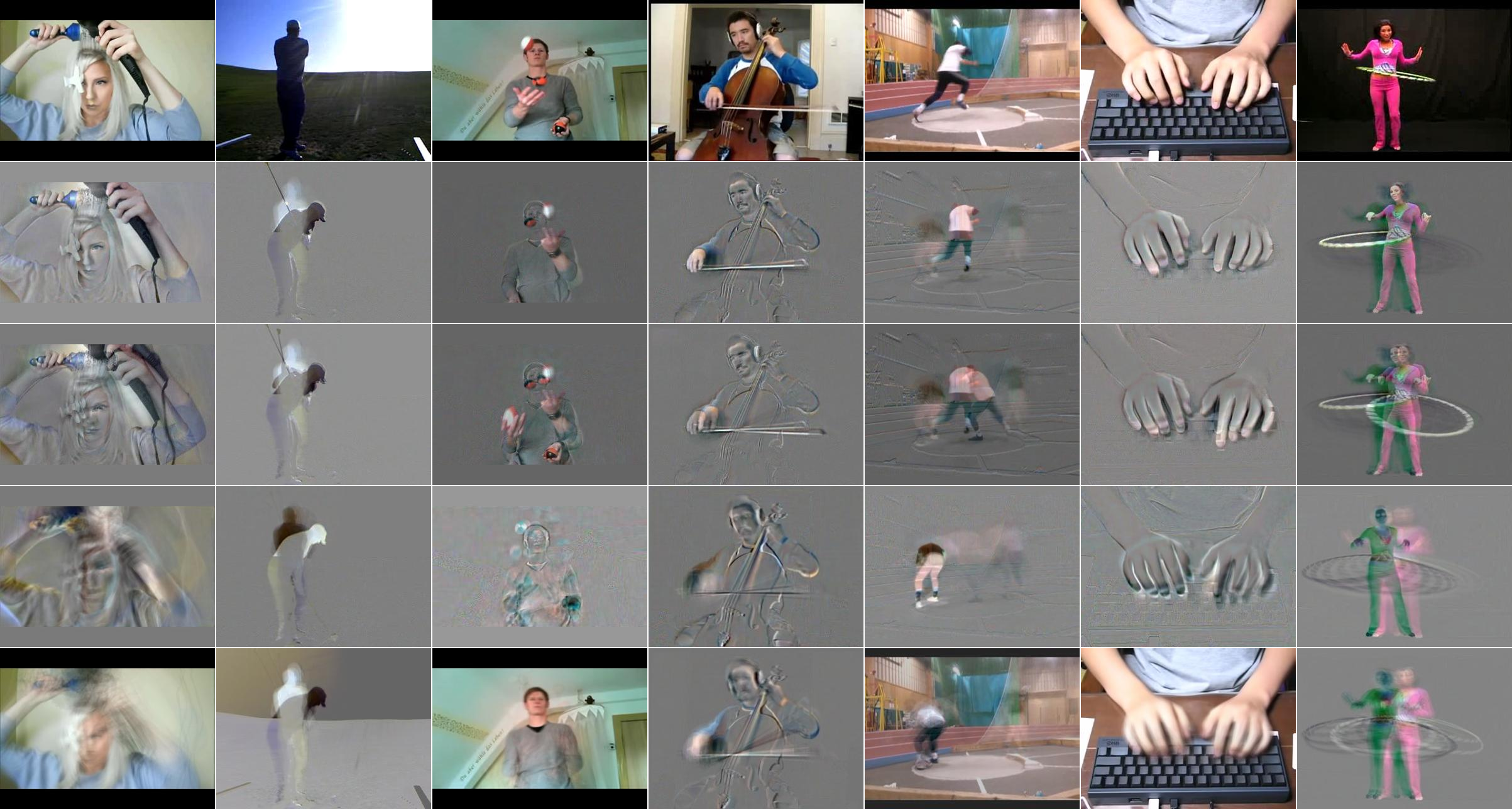}
	\caption{Representation of videos showing the actions (from left to right) blow-drying hair, playing golf, juggling, playing cello, shot putting, typing, hula hooping. Each row corresponds to an image type. From top to bottom: RGB image, flow profile image ($r=1$), flow profile image ($r=2$), dynamic image, and eigen image (first eigen evolution function).}
	\label{fig:flow_profile_images}
\end{figure*}

\section{Experiments}
\label{sec:experiments}

To demonstrate the capabilities of flow profile images, we compare them with dynamic images and eigen images for the task of action recognition.

\subsection{Dataset}
For evaluation, we use the well known action recognition dataset UCF101 \cite{DBLP:journals/corr/abs-1212-0402}.
It contains $13320$ videos for altogether $101$ action categories.
Every action category comprises $25$ groups, each consisting of four to seven video clips.
The videos in each group share special features like the acting person or the background.
All clips are user-uploaded videos from the Internet and can therefore be considered as realistic videos, captured in unconstrained environments.
The videos vary in length but are trimmed around the respective action.

\subsection{Setup}

We compute for each video of UCF101 a dynamic image, using the approximate method proposed in \cite{8094009}, an eigen image, and different flow profile images, varying $r$ from one to five.
Since, to the best of our knowledge, the official code for eigen evolution pooling has not been released, we use our own reimplemented version, following the approach described in \cite{DBLP:journals/corr/abs-1708-05465}. 
We compute the eigen images with the first eigen evolution function, since the authors report the highest accuracy for it among three tested evolution functions (evaluated using globally pooled RGB images on the first split of UCF101).
Moreover, mean and max images are created for each video by simply mean or max pooling all of its frames.
These two trivial image types serve as benchmarks in our experiments to which the other types are compared.

After generating all images, we fine-tune the BVLC reference model CaffeNet \cite{DBLP:conf/mm/JiaSDKLGGD14}, pre-trained on ImageNet ILSVRC 2012 \cite{DBLP:journals/ijcv/RussakovskyDSKS15}, for each image type on UCF101.
For the sake of comparison, we use the exact same fine-tuning routine for each of them.
More specifically, each image type is fine-tuned for roughly $160$ epochs, with the learning rate being decreased by a factor of ten roughly every $30$ epochs.
The model is evaluated approximately every five epochs and the respectively highest observed accuracy is reported.
During training, the images are randomly flipped and cropped, whereas in test phase, the center crop is used.
Even though there are improved Conv\-Net architectures, we decide to use CaffeNet since it enables an efficient training process.
To obtain state-of-the-art results, it is necessary to rely on a considerably stronger Conv\-Net and train it on several types of images. 
Bilen et al. \cite{8094009}, for example, train ResNeXt \cite{8100117} models on four image types, namely static images, optical flow images, dynamic images, and dynamic optical flow images.

\subsection{Results and Discussion}

\begin{figure}
	\centering
	\def\svgwidth{\columnwidth}
	\includegraphics[width=0.45\textwidth]{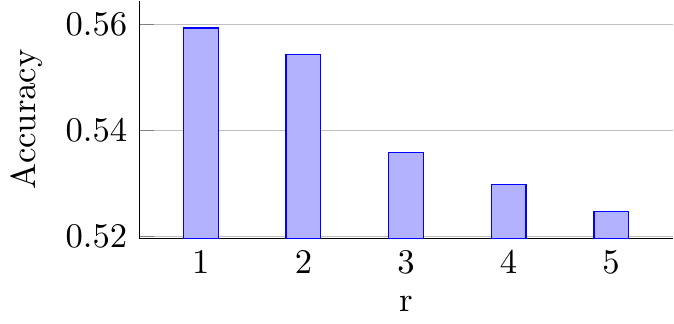}
	\vspace{-10pt}
	\caption{Action recognition accuracy of flow profile images by parameter $r$. Evaluated on first split of UCF101.}
	\label{fig:accuracy_by_r}
\end{figure}

\begin{table}
\centering
\begin{tabular}{|l|c|c|c|c|c|}
\hline 
\multirow{2}{*}{Image Type} & \multicolumn{4}{|c|}{Accuracy (\%)} \\ \cline{2-5} & {1} & {2} & {3} & {$\varnothing$}  \\
\hline
Flow Profile Image & 55.9 & 57.7 & 57.1 & 56.9 \\ \hline
Dynamic Image & 53.9 & 55.9 & 53.3 & 54.4 \\ \hline
Eigen Image & 52.8 & 51.3 & 52.0 & 52.1 \\ \hline
Mean Image & 51.6 & 50.9 & 49.7 & 50.8 \\ \hline
Max Image & 45.7 & 43.7 & 45.1 & 44.8 \\ \hline
\end{tabular} 
\caption{Action classification accuracy of different image types on UCF101.}
\label{tab:image_type_comparsion}
\end{table}

In figure \ref{fig:accuracy_by_r}, the accuracy of flow profile images is shown by parameter $r$, evaluated on the first split of UCF101.
When increasing $r$, a decrease in the action recognition performance is visible.
This might be attributed to the fact that the resulting images become overloaded when fusing too many frames with the same high flow energy score.
Conversely, flow profile images with $r=2$ suffer only a minor reduction in accuracy compared to flow profile images with $r=1$.
Concurrently, they contain in general more motion cues than images computed with $r=1$. 
Even though being less discriminative when evaluated alone, it would be interesting to see if they complement static RGB images better than flow profile images with $r=1$ do.

Table \ref{tab:image_type_comparsion} compares different image types in terms of action classification accuracy on UCF101. 
The evaluated flow profile images (with $r=1$) consistently obtain a higher accuracy on all three splits than both dynamic and eigen images.
Furthermore, images aggregated by simple mean or max pooling of all video frames perform constantly worse than images generated by the three more sophisticated pooling algorithms.
It is important to mention that Bilen et al. \cite{8094009} also evaluate their proposed (approximate) dynamic images on the first split of UCF101 using CaffeNet and report a higher accuracy, namely  $55.2 \%$ compared to $53.9 \%$ in our experiments.
Notably, also the accuracies achieved by mean and max images are reported to be higher.
Explanations for this difference can be attributed to various aspects of the approach, including a different preprocessing step or other hyperparameters.
When using exactly their routine, we would expect that the accuracies of eigen and flow profile images would improve as well.
Nevertheless, even when compared to the reported higher values, our flow profile images provide superior results.

\section{Conclusion}
In this paper, we have proposed a novel video representation, so-called flow profile image, that can be used in video classification tasks like activity recognition. The construction of flow profile images is computationally not complex and easy to implement.
Only RGB images and optical flow fields are required for their computation, both of which are standard in popular video classification architectures. Our conducted experiments show that these images can improve the action classification accuracy. As future work, one might consider flow profile images in other video processing applications such as gait recognition, video synopsis, and video summarization.


\bibliographystyle{splncs}
\bibliography{opticalflowimage}
\end{document}